# Identity-Centric Video Summarization via Hierarchical Fusion of Biometric, Appearance, and 3D Body Features[1]

Milad Mirjalili*, Enrique Alegre Gutiérrez, Eduardo Fidalgo Fernández, Víctor González Castro, Rocío Alaiz Rodríguez, Manuel Castejón Limas

*Group for Vision and Intelligent Systems, Institute for Research and Innovation in Engineering (I4), Universidad de León, 24071, León, Spain*

## Abstract

This work presents a video summarization algorithm based on multi-object tracking and person re-identification. We integrate facial embeddings, 3D body-shape features, and visual appearance into a unified tracking framework. These representations enable hierarchical identity assignment and tracking through bi-directional anchoring, which robustly recovers trajectories under severe occlusion or low visual quality. From these stable trajectories, we generate a compact set of summaries for each identity. We select keyframes using a multi-factor weighting scheme that optimizes biometric clarity, social interaction, and motion dynamics, while Adaptive Non-Maximum Suppression ensures temporal diversity. Evaluation on a custom dataset demonstrates tracking stability, achieving an IDF1 of 97.89% and a MOTA of 95.79%. Compared to Top-K selection, our algorithm also increases visual diversity by 146%, temporal coverage by 89%, and information retrievability by 3.5%.

***Keywords:*** Video summarization, Multi-object tracking, Biometric feature fusion, Human-centric perception, Computer vision, Deep learning.

[1] This document is an extended English version of the paper originally published in Spanish as:

Mirjalili, M., Alegre, E., Fidalgo, E., González-Castro, V., Alaiz-Rodríguez, R., Castejón-Limas, M. (2026). Resumen de vídeo centrado en la identidad mediante fusión jerárquica de características biométricas, de apariencia y corporales 3D [Identity-Centric Video Summarization via Hierarchical Fusion of Biometric, Appearance, and 3D Body Features]. Revista Iberoamericana de Automática e Informática Industrial (RIAI), [VOLUME], [PAGES]. https://doi.org/[DOI]

When citing this work, please cite the original published version above.

* Corresponding author: mmir@unileon.es

## 1. Introduction

The growing availability of cameras and recording devices generates an enormous amount of video data worldwide every day, far exceeding human capacity for manual review and analysis (Meena et al., 2023). This has created a pressing need to automate the creation of concise video summaries that capture the main events and individuals present in long recordings (Argaw et al., 2024; Li et al., 2023). Video summarization (VS) reduces extensive footage to its most essential parts in a structured and compact format (Alaa et al., 2024; Sultani et al., 2018). VS has clear potential in domains such as surveillance, where recordings are simply stored (Tank, 2023) and only a minimal fraction of the footage is actually relevant (Danesh Pazho et al., 2023). A typical surveillance feed captures hours of normal, everyday behavior, with only a small fraction containing anomalous activity or footage involving specific individuals of interest. Manual review of this footage is therefore not only laborious but also prone to error due to operator fatigue, which creates a critical need for automated systems reliable enough to detect meaningful frames and present them in a structured, compact form. VS is also used to generate content for social media (Narwal et al., 2025), films, or sports, adding value (Wistia, 2025) through short, content-rich videos (Gujar, 2024).

The quality of person-centric summarization depends in part on the Multi-Object Tracking (MOT) algorithm employed (Mirjalili et al., 2025). These algorithms aim to detect and track individuals (S. Li et al., 2025) despite occlusions and changes in appearance (Sharghi et al., 2017; Zhang et al., 2022). However, their performance can degrade in crowded and dynamic environments, where challenges such as occlusion, identity switches, and appearance variation persist.

To address this limitation, we propose a method that combines person tracking with identity-centric video summarization, integrating facial biometrics, appearance, and 3D body shape. This combination helps maintain identity consistency even when one feature becomes unreliable (e.g., during severe facial occlusion). We additionally provide a comprehensive evaluation of both the system's tracking performance using standard metrics and the representativeness of the generated summaries.

The main contributions of this work are:

- **Robust feature integration:** We combine AdaFace (facial biometrics) (Kim et al., 2022), TransReID (visual appearance) (He et al., 2021), and SMPL parameters (3D body shape) (Goel et al., 2023), reducing identification ambiguity in scenarios with occlusion or visual similarity.
- **Hierarchical association framework:** We propose a two-stage tracking strategy that prioritizes high-confidence facial anchors, followed by a matching stage that recovers occluded subjects. This strategy significantly reduces identity switches by ensuring that tracklets are initialized only from biometrically verified detections.
- **Dynamic multi-objective cost function:** We define a custom association cost function that weighs facial, appearance, body-structure, and spatial costs based on feature availability. This allows the tracker to alternate between matching modes without manual tuning.
- **Persistent gallery and trajectory recovery:** We maintain a stable representation of each identity and refine trajectories accordingly. This allows the system to close temporal gaps caused by long-term occlusions and ensures narrative continuity in the final summary.
- **Identity-centric summarization:** We generate individual summaries through multi-factor weighting that optimizes biometric clarity, social interaction, and motion dynamics. We additionally introduce Adaptive Non-Maximum Suppression (A-NMS) to improve temporal coverage and produce summaries that balance visual quality and diversity.

## 2. Related work

VS and Multiple Object Tracking (MOT) are broad research fields with diverse methodologies (Argaw et al., 2024; Lee et al., 2025; Saini et al., 2023). In this section, we first review the main categories and common architectures of VS. We then examine tracking algorithms proposed in the literature. Finally, we analyze prior work that attempts to link summarization with tracking, highlighting the limitations that our work aims to address.

### 2.1. Video summarization paradigms

VS algorithms can be supervised, unsupervised, or weakly supervised (H. Li et al., 2025).

***Supervised learning:*** Model training is guided by frame importance (Apostolidis et al., 2021c; Hsu et al., 2023; Jiang and Lan, 2025; Qaroush et al., 2025). These methods are limited by the scarcity of annotated data and the subjectivity inherent to the annotation process (Meena et al., 2023).

***Unsupervised learning:*** In the absence of annotations, techniques such as Generative Adversarial Networks (GANs) or Reinforcement Learning (RL) are used (Li et al., 2024; Liu et al., 2022; Yuan and Zhang, 2023). In the former, a Generator (G) predicts frame importance while a Discriminator (D) attempts to distinguish the generated summary from a reconstruction of the original video (Apostolidis et al., 2021a). In the latter, agents learn to select frames that maximize a reward function based on representativeness and diversity (Li and Yang, 2021).

***Weakly supervised learning:*** These methods use annotations such as video tags or titles instead of frame-level labels (Ramos et al., 2023). For example, textual descriptors can be used to align video segments with semantic concepts, and the model learns frame importance based on its relevance to the overall content of the video (Argaw et al., 2024).

### 2.2. Query-based and multi-modal summarization

To mitigate the subjectivity of generic summarization (Narwal et al., 2022), query-based approaches adapt the output using text descriptions or reference images (Bhute et al., 2025; Wu et al., 2022). These approaches rely on integrating diverse features and semantically relating different types of data. One approach aligns natural language queries (e.g., "show me the goal") with visual content (Narasimhan et al., 2021; Xiao et al., 2020). This typically involves using separate encoders for visual frames and text, which are later merged into a shared latent space, or using cross-modal attention (Argaw et al., 2024). Another approach is to generate more context-aware summaries (Zhao et al., 2023) by incorporating audio information.

However, few works explicitly focus on identity-centric summarization, where the query corresponds to the biometric or morphological identity of a specific individual (Mirjalili et al., 2025). While early attempts relied on simple object or face detection filters, extracting only the frames where a face was detected with high confidence (Biswas et al., 2021), these methods lack the ability to differentiate specific individuals or maintain narrative continuity. Our work addresses this gap with a multi-cue tracking method that serves as a query mechanism for filtering and summarizing footage based on individual identities.

### 2.3. Spatiotemporal architectures

Video processing involves capturing both spatial and temporal information (Apostolidis et al., 2021b). Early deep learning approaches used 3D CNNs to extract spatiotemporal features, or hybrid architectures combining 2D CNNs (for spatial features) with RNNs, such as Long Short-Term Memory (LSTM) or Gated Recurrent Units (GRU), to model temporal dependencies (Alomar et al., 2025). More recently, Vision Transformers (ViT) and multimodal Transformers have become state-of-the-art, using self-attention to capture long-range dependencies across the entire video sequence, often outperforming RNNs in modeling long-range dependencies (Tang et al., 2025).

### 2.4. Multi-Object Tracking (MOT) and re-identification

In MOT, the dominant paradigm is tracking-by-detection, in which targets are detected in each frame and subsequently associated into continuous trajectories (Guan et al., 2025). Most deep learning MOT systems are based on this paradigm (Hassan et al., 2023; S. Li et al., 2025).

Algorithms such as SORT (Simple Online and Realtime Tracking) use a Kalman filter to locate objects and the Hungarian algorithm for association, resulting in a simple motion-based tracking framework (Bewley et al., 2016). DeepSORT adds CNN-based appearance embeddings to help preserve identity through occlusions and abrupt motion changes (Wojke et al., 2017). ByteTrack improves tracking by leveraging both high- and low-confidence detections to recover occluded objects and maintain continuity (Zhang et al., 2022). Methods such as DeepOCSORT employ a fusion strategy (Maggiolino et al., 2023) that weighs motion and appearance, prioritizing visual identity when motion is erratic, or spatial overlap when visual features are ambiguous.

Early deep re-identification (ReID) methods relied on CNNs such as ResNet50 (He et al., 2016) to compute appearance descriptors that support identity consistency across frames (Wojke et al., 2017). More recently, they rely on Transformers such as TransReID to capture fine-grained, part-aware visual features (e.g., clothing, accessories) with stronger discriminative power than standard global CNN embeddings (He et al., 2021). ReID performance can degrade for individuals wearing similar clothing or under severe occlusion. Human Mesh Recovery (HMR), which estimates the parameters of a parametric body model (SMPL) (Goel et al., 2023), can provide additional identification cues when visual features are unreliable.

To synthesize this review and establish the context of our contribution, Table 1 presents a structural comparison of standard MOT trackers against our proposal. As the analysis shows, the considered tracking methods primarily rely on kinematic or 2D visual associations and do not explicitly incorporate 3D structural or biometric constraints. Our proposal addresses this gap by incorporating biometric and 3D structural cues for long-term tracking.

### 2.5. Linking tracking and summarization

Existing evaluation protocols and datasets indicate that MOT and VS are often addressed separately (Dendorfer et al., 2021; Otani et al., 2019). VS selects keyframes by optimizing coverage and diversity on datasets such as SumMe and TVSum (Gygli et al., 2014; Yale Song et al., 2015). MOT maintains the trajectory consistency of individual targets on datasets such as MOTChallenge (Dendorfer et al., 2021). Despite this separation, several studies have attempted to combine these two fields to produce human-centric summaries. Video Synopsis extracts moving objects (tubes) and rearranges them temporally to display multiple events simultaneously, condensing hours of footage into minutes (Baskurt and Samet, 2019; Zhang et al., 2024). Other methods employ tube-level scoring, retaining the segments with the highest representativeness (Shoitan et al., 2023). However, these methods may depend on hand-crafted energy functions or generic metrics that do not explicitly maintain identity consistency in complex environments (Li et al., 2018).

**Table 1: Structural comparison of standard MOT tracking methodologies against our proposal.**

| Method | Appearance extraction | Association mechanism | Main strengths | Critical limitations |
|---|---|---|---|---|
| SORT (Bewley et al., 2016) | None. Purely based on linear kinematics. | Kalman filter + Hungarian algorithm. Assumes linear, constant-velocity motion for spatial association (IoU). | Extremely fast and efficient. | Performance degrades substantially during prolonged occlusions. |
| DeepSORT (Wojke et al., 2017) | Yes. Offline CNN network for appearance metrics. | Adds a cascaded matching strategy and Mahalanobis distance to the SORT motion model. | Reduces ID switches by roughly 45% relative to SORT; handles short occlusions. | Slower; trajectories can jump to false positives generated by static background geometry. |
| ByteTrack (Zhang et al., 2022) | Conditional. Only on high-confidence detections. | BYTE association: associates all boxes, using the low-score ones to recover occluded objects through pure IoU. | Recovers heavily occluded objects smoothly. | For low-confidence boxes, visual appearance is unusable, forcing total reliance on kinematics (IoU). |
| Deep OC-SORT (Maggiolino et al., 2023) | Yes (adaptive). Appearance weighting is adjusted dynamically according to confidence. | Improves the SORT linear motion model by incorporating Momentum Recovery and Camera Motion Compensation (CMC) to associate non-linear trajectories under occlusion. | Strong performance in complex dynamics; very robust to non-linear motion. | High computational and architectural complexity; requires tuning multiple modules and operates on 2D image-space cues. |
| Our proposal | Yes (multiple features). Facial biometrics (AdaFace), visual appearance (TransReID), and 3D body shape (SMPL). | Bi-directional hierarchical tracking: a dynamic cost function adaptively integrates spatial distance, facial error, visual similarity, and 3D geometric consistency. | Viewpoint invariance; robust recovery after 180° turns and severe occlusions by validating 3D structure. Optimized for the post-processing stage in high-precision video summary extraction. | Dependence on a simple linear kinematic model; higher computational cost due to the parallel extraction of multiple networks. |

In a manner analogous to tracking, Table 2 contrasts summary generation across the different approaches found in the literature and our proposal. An analysis of these methodologies reveals that methods relying on isolated 2D attributes or purely kinematic cues remain fragile under complex interactions. Our approach overcomes these paradigms by using biometric and structural validation to extract a narrative summary centered exclusively on the individual of interest.

In our previous work (Mirjalili et al., 2025), we linked person-centric tracking with video summarization, combining intermediate YOLO layers (Varghese and M., 2024), 2D pose landmarks to extract structural motion, and ArcFace embeddings for biometric identity (Deng et al., 2019). Rather than performing strict frame-by-frame tracking, we used a clustering-based approach to group detections into unique identity clusters.

**Table 2: Methodological comparison of the main summarization approaches in the literature against our proposal.**

| Approach in the literature | Main selection method | Identity handling | Main strengths | Limitations |
|---|---|---|---|---|
| Traditional video summarization (Jiang and Lan, 2025; Qaroush et al., 2025; Li et al., 2024) | Keyframe extraction that skips redundant segments based on global scene variance. | None. Depends on global visual features of the whole frame, without associating subjects. | Computationally fast; maintains a real timeline without artificial overlaps. | May disrupt action continuity; does not allow the summary to focus on a specific person. |
| Filter-based summarization, face/body (Biswas et al., 2021) | Filtering through isolated local thresholds (e.g., high-confidence frontal face detection). | Weak. Relies purely on the point-wise visibility of 2D attributes in isolated frames. | Allows some personalization; reduces duration by showing the subject only when visible frontally. | Performance degrades under occlusion, low resolution, or when the subject turns their back to the camera. |
| Video synopsis, motion tubes (Zhang et al., 2024; Shoitan et al., 2023) | Extraction and temporal shifting of moving objects (“tubes”) to display them simultaneously over a static background. | Moderate. Maintains short trajectories based on pure kinematics or background subtraction. | Can substantially reduce viewing time by displaying sequential events in parallel. | May produce visually crowded scenes, separate people who are moving together, and lose identity consistency during dense crossings. |
| Our proposal | Keyframe extraction for each detected ID, guided by Adaptive Non-Maximum Suppression (A-NMS). | Designed for robust identity association. Dynamic cost function anchored in biometrics, visual features, and 3D structure. | Query-based generation. Its selection criteria let the summary focus on activity level, social interaction, and visual resolution. | An error in ID assignment can introduce different people into a subject's summary; higher computational load. |

While this previous work demonstrated the feasibility of identity-based summarization, it highlighted several critical limitations that we now aim to address:

- **Feature specificity:** Intermediate YOLO layers provide less discriminative identity features than dedicated ReID architectures for distinguishing individuals with similar appearance.
- **2D versus 3D structure:** Relying on 2D pose landmarks makes the system susceptible to viewpoint variation. A person's 2D skeleton can differ substantially between side and frontal views, potentially contributing to identity switches. In contrast, 3D body shape provides a view-invariant biometric signature that remains stable regardless of the subject's orientation relative to the camera.
- **Biometric robustness:** Although ArcFace is effective, its fixed angular margin is independent of image quality, which can limit its effectiveness on blurry or noisy inputs. We replace ArcFace with AdaFace, whose adaptive margin is based on image quality and is designed to improve recognition for low-resolution faces commonly encountered in surveillance footage (Kim et al., 2022).

Table 3 summarizes the methodological changes from our preliminary work and their intended effects, summarizing the improvements introduced in feature extraction, temporal association, and summary generation. Notably, while our previous approach generated global, statistical summaries, the robustness of our new architecture enables a query-based summarization paradigm, generating independent, high-quality visual narratives for each individual.

**Table 3: Architectural comparison detailing the methodological evolution from the preliminary work to the current proposal.**

| Algorithm component | Previous work (Mirjalili et al., 2025) | Proposed method | Impact / solution provided |
|---|---|---|---|
| Facial biometrics | ArcFace. Applies a fixed angular margin regardless of image quality. | AdaFace, an adaptive margin based on image quality and resolution. | Improves detection and recognition under occlusion and low-resolution conditions. |
| Visual appearance | Intermediate YOLO layers. Extract very general features. | TransReID (Vision Transformer). | Self-attention captures fine-grained detail (clothing, accessories), surpassing global CNN descriptors. |
| Body structure | 2D keypoints. | 4D-Humans (3D SMPL parameters). | Viewpoint invariance. The 2D skeleton fails when the subject turns, while the 3D signature remains stable. |
| Association logic | Direct, static clustering. | Hierarchical, bi-directional anchoring with a dynamic cost function. | Reweights association dynamically when the face is hidden, relying instead on body shape and motion. |
| Summary selection | Global, statistical summary. | Individual per-ID summary guided by a multi-factor function and A-NMS. | Moves from a generic statistical summary to an independent narrative for each person, optimizing biometric clarity, activity, and interactions. |

# 3. Proposed algorithm

This work extends the framework presented in (Mirjalili et al., 2025) by introducing significant improvements: (i) features that are more robust to occlusion and pose variation, (ii) a hierarchical association scheme for stable tracking, and (iii) an automated, identity-centric summarization method. Our proposal prioritizes the extraction of semantically informative keyframes for social media content and identity-centric analysis. To this end, we adopt a feature-centric architecture rather than a motion-centric one. This choice is intended to reduce unnecessary computation while maintaining the visual quality required in the summaries. The overview of the method is illustrated in Figure 1.

## 3.1. Person localization and detection

The first stage requires precise localization of the individuals in the scene. Given an input video sequence $V = \{I_1, I_2, \dots, I_T\}$, where $I_i$ denotes the i-th frame, we employ an object detection model to identify people. The output of this module is a set of bounding boxes $B_t = \{b_1, b_2, \dots, b_n\}$ for the individuals detected in each frame, which serves as input for feature extraction.

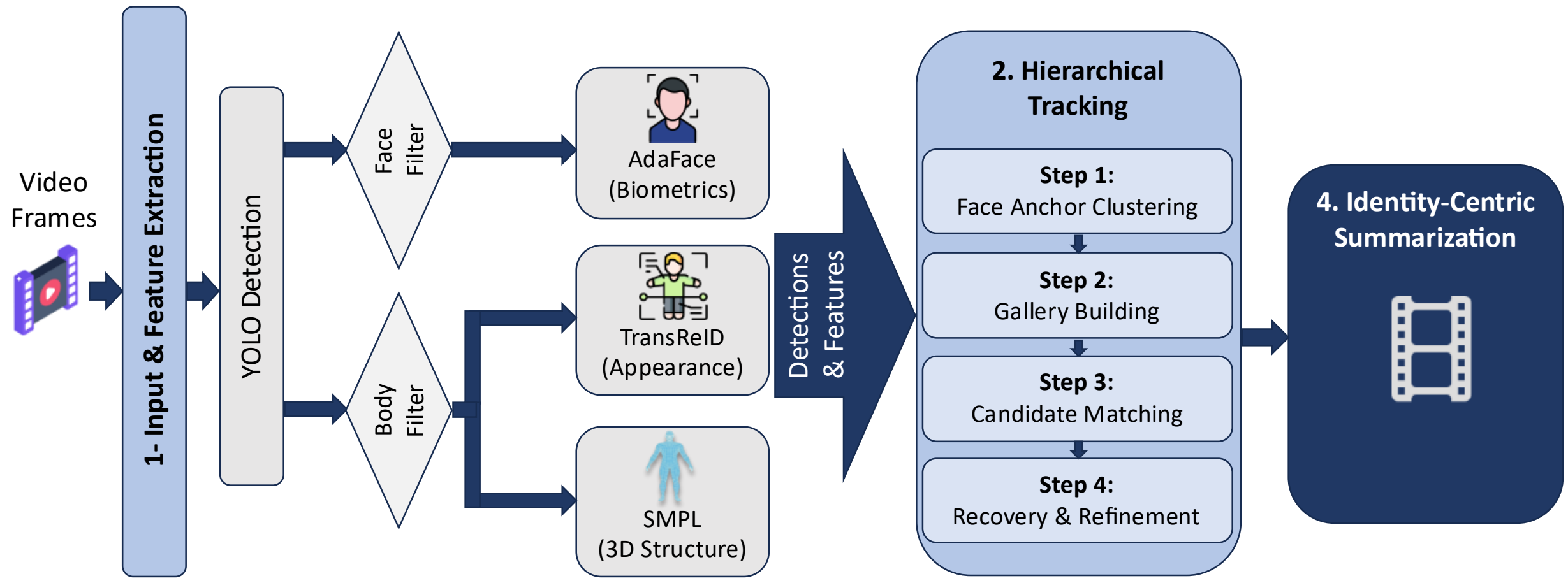

**Figure 1: Overview of the proposed hierarchical tracking and summarization framework. The pipeline begins by extracting features from video frames, where quality filters are applied before computing the AdaFace (biometrics), TransReID (visual appearance), and SMPL (3D body shape) embeddings. The tracking process consists of four hierarchical steps: (1) Face Anchor Clustering uses DBSCAN on high-quality faces to find unique identities; (2) Gallery Construction establishes a robust profile for each cluster; (3) Candidate Matching links the remaining detections (e.g., occluded or back-turned) to these profiles; and (4) Recovery and Refinement handles unmatched tracklets and smooths the trajectories. Finally, the system generates an Identity-centric summary containing selected keyframes and metadata for each person.**

## 3.2. Feature extraction

For each detected individual, we extract features representing facial biometrics, visual appearance, and 3D body shape. For computational efficiency, frames without detected people are skipped. We use two stages to verify identity. First, we apply the Sample and Computation Redistribution Face Detector (SCRFD) (Guo et al., 2021), which is designed to provide high recall for small and occluded faces commonly encountered in surveillance footage. This step yields a facial detection score that allows low-confidence detections to be discarded. The face image is then aligned using facial landmarks and fed into AdaFace to extract the identity embedding $f_{face}$, using a loss function whose margin is adaptive to image quality (Kim et al., 2022).

***Semantic pose filtering:*** To apply computationally expensive body feature extractors only to meaningful data, we implement a validity filter leveraging the YOLO detector's pose estimation output. We analyze the confidence scores and spatial distribution of the predicted skeletal keypoints to assess structural integrity. This module acts as a gating mechanism, distinguishing valid human poses, including challenging back views or headless torsos, from detector artifacts such as disembodied hands or background clutter. By filtering out detections with sparse or low-confidence keypoints, this step reduces redundant computation and prevents the re-identification gallery from being corrupted with noisy embeddings. For detections that pass this filter, we extract the remaining body-centric features.

For subjects with occluded faces or facing away from the camera, we employ the Vision Transformer-based TransReID (He et al., 2021). Unlike CNN-based approaches, which focus on global texture, this algorithm uses self-attention to capture fine-grained details (e.g., shoes, backpacks, patterns). We extract a robust appearance feature map $f_{app}$ that encodes the subject's clothing and overall body semantics. This design helps maintain the trajectory during short-term facial occlusions.

Finally, to improve robustness to viewpoint variation, we use 4D Humans (HMR 2.0) to reconstruct the 3D human mesh from 2D images (Goel et al., 2023). Specifically, we extract the SMPL β parameters (shape parameters), denoted $f_{shape}$. These parameters represent aspects of the person's physical structure, such as height, limb proportions, and girth, which are relatively stable across camera viewpoints. $f_{shape}$ acts as a geometric consistency constraint, serving as a negative filter to prevent matching individuals with very different morphologies even if their clothing looks similar.

### 3.3. Hierarchical tracking

To maintain consistent identities in challenging scenarios, such as occlusions, 180° turns, and side-profile views, we propose a multi-stage, bi-directional association strategy that uses high-confidence biometric anchors.

***Step 1: Face anchor clustering:*** We identify detections with high-quality facial features across the sequence. Rather than relying on a local threshold, we aggregate all valid face embeddings $E_{face}$ and apply DBSCAN (Ester et al., 1996) to identify unique identities globally.

***Step 2: Gallery construction:*** For each cluster $C_k$, we establish a gallery entry $G_k$ containing:

- The centroid of the facial embeddings ($\mu_{face}$).
- A set of K distinct appearance embeddings ($E_{app}$), sampled from frames where the subject's face was clearly visible. To ensure diversity, we keep the K most distinctive embeddings (e.g., front, back, side) based on cosine dissimilarity, minimizing feature redundancy.
- The median of the body shape parameters ($\beta_{median}$), as a robust structural signature.

Detections within these clusters serve as high-confidence anchors for subsequent association.

***Step 3: Candidate matching:*** We link the remaining detections, including those with occluded or non-frontal faces, to the established anchors using a bidirectional matching strategy. We define a dynamic cost function $C(d, g)$, shown in Eq. (1), to compute the dissimilarity between a candidate detection $d$ and a tracked identity $g$,

$$C(d, g) = \frac{w_a D_a + w_s D_s + w_l D_l + \alpha_f . w_f D_f}{w_a + w_s + w_l + \alpha_f . w_f} \rho(\Delta t), \quad (1)$$

where:

- $D_a$: cosine distance between the candidate's ReID embedding and the closest match within gallery g.
- $D_s$: Euclidean distance between the SMPL shape parameters β.
- $D_l$: spatial distance derived from a linear velocity prediction model.
- $D_f$: cosine distance for facial embeddings, active only for visible faces (setting $\alpha_f = 1$).
- $\rho(\Delta t)$: a penalty that increases as the temporal gap ($\Delta t$) between frames grows. Empirically, this function is defined as $\rho(\Delta t) = 1.0 + \alpha \Delta t$, with $\alpha = 0.02$. This linearly scales the association cost, gradually reducing the probability of linking subjects that have been absent from the scene for extended periods.

The dynamic weighting is governed by the indicator function $\alpha_f$. When a face is detected ($\alpha_f = 1$), the high-confidence facial score is included in the weighted average. When the subject turns away or the face is occluded ($\alpha_f = 0$), the facial term is dropped, and the weights are automatically redistributed to prioritize $D_a$, $D_s$, and $D_l$, ensuring robust tracking based on body features and motion.

***Step 4: Unmatched recovery and post-processing:*** We process unidentified detections without reliable facial features by clustering their appearance embeddings with DBSCAN to identify additional tracklets. Fragmented tracklets are then merged and trajectories are smoothed. This step unifies fragmented identities by analyzing the temporal and visual proximity of disconnected segments, closing short occlusion gaps, and improving spatiotemporal continuity.

### 3.4. Summary selection

We formulate the summarization of long-duration sequences as an optimization problem: for each unique identity ($ID_i$), we seek a subset of keyframes that maximizes semantic information while minimizing temporal redundancy. To this end, we introduce a frame-level importance score, prioritizing frames according to biometric quality, social context, resolution, and motion dynamics.

***Importance scoring function:*** For a target person P at frame t, the total importance $S(P, t)$ is defined from four normalized factors, as shown in Eq. (2):

$$S(P, t) = w_1 Q_{face} + w_2 I_{social} + w_3 Q_{appear} + w_4 M_{action} \quad (2)$$

where:

- $Q_{face}$: scores confidence in face detection, favoring frames in which the subject's face is clearly visible and frontal.
- $I_{social}$: frames containing social interactions are semantically more relevant. We measure interaction across all concurrent detections and select the maximum value corresponding to the nearest neighbor, as shown in Eq. (3):

$$I_{social} = \exp\left(-\frac{\left\|c_p - c_j\right\|_2^{\,2}}{2\sigma^2}\right), \quad (3)$$

where $c_p$ and $c_j$ are the centroids of the bounding boxes of the target and the nearest neighbor, and σ is a scale parameter. The Gaussian kernel naturally models human social zones, where proximity decays gradually rather than abruptly.

- $Q_{appear}$: the ratio between the bounding-box area and the frame area, or pixel density of the subject. High values prioritize close-ups with clearly visible detail over distant, low-resolution detections.
- $M_{action}$: the normalized velocity. Let $c_t$ and $c_{t-1}$ be the bounding-box centroids in the current and previous frames, and $\delta_{max}$ an empirical velocity limit. We define $M_{action}$ as shown in Eq. (4):

$$M_{action} = min\,\{\,1\,;\ \delta_{\max}^{-1} || c_t - c_{t-1} ||_2\,\} \quad (4)$$

Restricting the velocity to the $[0, 1]$ interval prevents very high-speed movements from dominating the weighted sum. This factor prioritizes frames in which the subject is performing an action (e.g., walking) over stationary frames, adding dynamic context to the summary.

***Temporal diversity via adaptive NMS:*** Selecting frames purely by score tends to produce redundancy, with all selected frames clustering around a single high-quality moment. To ensure temporal coverage, we apply Adaptive Non-Maximum Suppression (A-NMS). Selecting a keyframe at time t with the highest score suppresses the score of all neighboring frames within a temporal window $[t - \Delta,\ t + \Delta]$ (with $\Delta =$ 2 seconds). This forces the selection algorithm to identify the next most informative frame from a distinct time segment, ensuring that the storyline covers the entire duration of the track.

***Summary generation:*** For each processed identity, we generate a structured summary package containing:

- **Representative thumbnail:** the single frame with maximal $Q_{face}$ , serving as the biometric profile image.
- **Chronological timeline:** a sequence of $K$ keyframes sorted in temporal order that represents the subject's narrative. The timeline can be easily personalized by adjusting the weighting terms; for instance, a user may prioritize social interactions for security investigations, or motion dynamics for sports analytics, by adjusting the corresponding coefficients in the importance function.

- **Semantic metadata:** a profile card that includes the specific ID, total track duration, a list of concurrent identities derived from the associated detections, and the dominant motion state derived from the average $M_{action}$.

## 4. Experiments and results

Our experiments focused on two key aspects: tracking accuracy and the representative quality of the generated summaries. We evaluated the tracking algorithm, feature set, and post-processing pipeline through an incremental ablation study.

### 4.1. Experimental setup

***Datasets:*** We used Custom-YT, our own dataset extended from Mirjalili et al. (2025) with more processed frames and ground-truth bounding boxes. For reproducibility, it is publicly available in the repository of the Group for Vision and Intelligent Systems (GVIS)[2]. This evaluation was not intended to compete with massive tracking benchmarks, but rather to assess the improvements of this proposal in scenarios representative of the intended application. We therefore use a dataset that presents variability consistent with social media content. Large-scale evaluation is left for future work.

Custom-YT is diverse in its challenges, such as lighting conditions, camera angles, and backgrounds, and is representative of social media content, acting as a demanding "micro-benchmark" for the algorithm. The sequences include: (i) frames with degraded facial quality, simulating video surveillance; (ii) severe occlusions or subjects with their backs to the camera, to evaluate re-identification through visual appearance and 3D shape; (iii) lighting changes; and (iv) multiple subjects disappearing for long periods, to evaluate the consistency of long-term matching. Figure 2 shows examples from the dataset that directly illustrate these visual challenges.

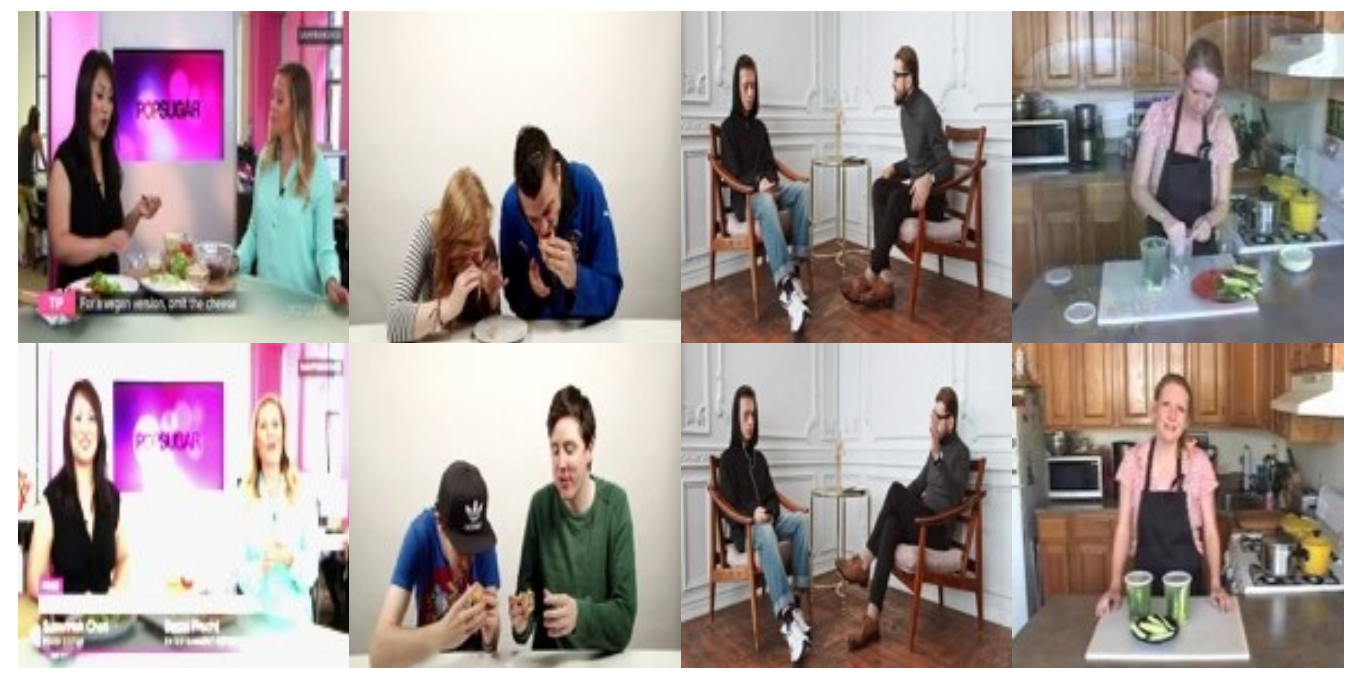

**Figure 2: Visual challenges in Custom-YT. From left to right: the first column illustrates subjects with their backs turned and lighting changes; the second, facial absence and prolonged disappearances used to evaluate long-term matching; the third, low resolution with degraded quality; and the fourth, variable poses and camera angles.**

Beyond visual complexity, this selection makes the dataset a rigorous stress test at the quantitative level. Table 4 provides the video statistics, trajectory duration, and facial availability, whose absence (16% of detections) requires the system to rely exclusively on body and appearance features to maintain tracking.

[2] https://gvis.unileon.es/datasets_main/

**Table 4: Statistics of the dataset used, including the number of videos, resolution range (in pixels), number of frames, number of unique identities (#IDs), average number of people per frame (density), track length (in frames), and facial availability, illustrating the complexity of the subjects to be tracked.**

| #Videos | Resolution range (px) | #Frames | #IDs | Density | Track length (min-max) | Facial availability (%) |
|---|---|---|---|---|---|---|
| 5 | 320×240 to 3840×2160 | 5664 | 39 | 1.93 | 2 - 1305 | 84.07 |

***Evaluation metrics:***

***Tracking:*** We report standard MOT metrics. The main one, Higher Order Tracking Accuracy (HOTA) (Luiten et al., 2021), jointly evaluates object detection quality and temporal association. We additionally report Multiple Object Tracking Accuracy (MOTA) (Bernardin and Stiefelhagen, 2008), which reflects detection accuracy considering false positives, false negatives, and identity errors. To evaluate long-term identity consistency, we report the Identification F-Score (IDF1) (Ristani et al., 2016). Finally, we measure the frequency of incorrect trajectory reassignments using the number of Identity Switches (ID Sw).

***Video summarization:*** We evaluated optimization success, information preservation, and spatiotemporal diversity. We compared these metrics against uniform temporal sampling and Top-K selection of the highest-quality frames, to analyze the trade-offs between raw quality and representative selection. To verify that high-quality observations are prioritized, we report the average AdaFace confidence and the coverage of the selected bounding boxes. We evaluated the representative power of the summary using a Retrievability metric. Keyframes are treated as a gallery and the remaining frames as queries, evaluating the average cosine similarity between queries and summaries. To avoid redundancy, we also measure the average pairwise cosine distance between ReID embeddings, as well as the temporal coverage.

***Implementation details:*** The critical hyperparameters of the architecture were tuned empirically. In the detection stage, the SCRFD confidence threshold was set to 0.50, while the YOLO threshold was set to 0.45. For the initial clustering of facial anchors, DBSCAN used a maximum distance (eps) of 0.40 with a minimum of 4 samples (min_samples). In the secondary ReID phase, a stricter DBSCAN was applied (eps = 0.25, min_samples = 3) to avoid erroneously merging distinct subjects wearing similar clothing. During tracking association, the cost function prioritizes spatial continuity, giving the highest weight to spatial proximity ($w_l$ = 1.5), followed by facial and visual consistency ($w_f$= 1.0, $w_a$ = 1.0), and using 3D body shape as a structural safety constraint ($w_s$= 0.8), all evaluated under a matching threshold of 0.70. Finally, for video summary selection, biometric clarity dominates the equation ($w_1$= 0.40), while the contextual factors, social interaction, visual resolution, and motion dynamics, contribute equally with a weight of 0.20 each ($w_2 = w_3 = w_4$ = 0.20).

## 4.2. Comparative baselines and ablation study

All experiments used the same base object detection architecture (YOLOv8 Pose) and preprocessing.

***Tracking algorithms:***

- **Global clustering:** Used in Mirjalili et al. (2025). Aggregates features from all detections into a single graph and applies DBSCAN for identity clustering.
- **Bi-directional anchoring:** Proposed in this work and described in Section 3, it uses anchor-based trajectory initialization, dynamic association costs, and bi-directional propagation.

***Feature sets:***

- **Legacy features:** The baseline set from our previous work, consisting of standard CNN-based appearance embeddings (intermediate YOLO layers), flattened 2D pose keypoints from the YOLO pose model, and ArcFace embeddings.
- **Enriched features:** The set proposed in this work, comprising TransReID (appearance), 4D-Humans SMPL (3D body structure), and AdaFace (biometrics).

***Post-processing:*** The final stage of the pipeline refines the trajectory, smoothing it by eliminating temporal gaps.

***Experimental configurations:*** We measure the contribution of each component by evaluating five tracking configurations:

- **Legacy baseline:** Global clustering with the legacy features already used in Mirjalili et al. (2025).
- **Feature upgrade:** Global clustering with enriched features.
- **Tracking upgrade:** Bi-directional anchoring with legacy features, to evaluate the impact of the new association strategy alone.
- **Proposed combination:** Bi-directional anchoring with enriched features, without post-processing.
- **Proposed full system:** Adds the post-processing module to the proposed combination.

## 4.3. Tracking results

In Table 5, the legacy baseline achieves considerable performance (HOTA 84.36%). However, it proves fragile under occlusion, since identity is not preserved in the intermediate YOLO layers and 2D pose keypoint estimation is unstable, which translates into identity switches (30 ID Sw) in the absence of faces.

The feature upgrade produces an immediate improvement in identity consistency, reducing identity switches to 9. This is due to AdaFace's facial embeddings and their robustness to low quality, together with the reduced ambiguity achieved by the dedicated ReID and 3D shape features in challenging scenes.

In the tracking upgrade, although MOTA and IDF1 showed slight fluctuations compared to the baseline, identity switches were reduced to only two. This suggests that the legacy features lack the discriminative capacity to fully exploit the potential of the algorithm, limiting gains in recall and precision.

The proposed combination improved performance, eliminating identity switches and increasing MOTA to 95.06% and IDF1 to 97.51%. By hierarchically integrating complementary feature spaces, ReID and 3D shape manage to maintain consistency despite the absence of faces.

The proposed full system produced an additional performance gain, increasing HOTA to 85.13% and MOTA to 95.79%. This result indicates that offline refinement recovers missed detections and smooths trajectories without introducing additional identity errors.

**Table 5: Comparative baselines and ablation study. We evaluate the incremental contribution of each component on our custom dataset (Custom-YT). The legacy baseline represents our previous approach using global clustering and legacy features. The proposed full system integrates the bi-directional anchor tracker, the new feature stack (AdaFace, TransReID, 4D-Humans), and the post-processing refinement module.**

| Method | Tracker | Features | Post-proc. | HOTA ↑ (%) | MOTA ↑ (%) | IDF1 ↑ (%) | ID Sw ↓ |
|---|---|---|---|---|---|---|---|
| Legacy baseline | Global clustering | Legacy | No | 84.36 | 94.55 | 97.00 | 30 |
| Feature upgrade | Global clustering | Enriched | No | 84.53 | 94.86 | 97.22 | 9 |
| Tracking upgrade | Bi-directional anchoring | Legacy | No | 84.46 | 92.90 | 96.33 | 2 |
| Proposed combination | Bi-directional anchoring | Enriched | No | 84.75 | 95.06 | 97.51 | **0** |
| Proposed full system | Bi-directional anchoring | Enriched | Yes | **85.13** | **95.79** | **97.89** | **0** |

## 4.4. Feature ablation study

To isolate the individual contributions of each modality within the enriched feature set, we performed a component-wise ablation study using the final model configuration. To ensure that the evaluation reflects the intrinsic discriminative power of the online tracking features rather than offline refinement, all post-processing is disabled. We also evaluated the impact on computational cost (frames per second, FPS) of each modality. We define three configurations for this analysis, whose results are presented in Table 6:

- **Face only:** Uses high-confidence facial anchors for association and discards faceless detections.
- **Face + shape:** Integrates 3D shape parameters to maintain tracking when faces are unavailable.
- **Face + shape + appearance (full model):** Adds appearance embeddings (TransReID) to resolve complex visual ambiguities.

Using only high-confidence facial anchors produces a highly precise baseline with zero identity switches. However, this configuration exhibits reduced recovery (MOTA 92.49%), since tracks terminate immediately when faces are occluded or subjects turn around, limiting overall trajectory coverage.

Integrating the 3D shape feature improved all accuracy metrics (HOTA +0.32, IDF1 +0.22, MOTA +0.39). This suggests that body shape provides critical discriminative signals in the absence of faces. Notably, this configuration produced a small increase in identity switches (2), reflecting improved recovery: whereas the face-only baseline discarded difficult trajectories, avoiding errors, Face + shape successfully closed ambiguous segments, occasionally at the cost of a switch, but keeping the trajectory alive for longer.

The full model achieves the highest overall performance, increasing MOTA to 95.06% and IDF1 to 97.51% while eliminating identity switches. Integrating appearance-based ReID features resolves the residual ambiguities left by shape matching alone, confirming that robust long-term identity preservation requires the joint exploitation of face, shape, and appearance.

***Computational cost analysis:*** Finally, we evaluated hardware requirements and runtimes. All inference was performed on a workstation equipped with an NVIDIA A100 GPU with 40 GB of VRAM. The system showed high memory efficiency, with a peak of 4.9 GB in the heaviest configuration.

Table 6 shows a clear trade-off between empirical accuracy and processing speed. The Face-only baseline is computationally very efficient, reaching 25.04 FPS. However, once body shape is processed with 4D-Humans, speed drops to 4.90 FPS. For the full model we obtain 4.65 FPS. As this minimal difference shows, appearance extraction through TransReID turns out to be extremely lightweight. This suggests that 3D mesh estimation is the primary computational bottleneck. Therefore, while the full architecture guarantees maximum accuracy for offline video summarization or forensic analysis applications, strict real-time deployment will require further optimization.

**Table 6: Feature set ablation study. We analyze the incremental contribution of each component and its computational impact.**

| Configuration | HOTA ↑ (%) | MOTA ↑ (%) | IDF1 ↑ (%) | ID Sw ↓ | Avg. FPS ↑ |
|---|---|---|---|---|---|
| Face only | 84.12 | 92.49 | 96.10 | **0** | **25.04** |
| Face + shape | 84.44 | 92.88 | 96.32 | 2 | 4.90 |
| Face + shape + appearance | **84.75** | **95.06** | **97.51** | **0** | 4.65 |

## 4.5. Quantitative analysis of the summary

Table 7 shows that the quantitative evaluation reveals distinct performance trade-offs across the three selection strategies. The Top-K baseline, which prioritizes frames with the highest facial scores, obtained the highest biometric confidence and resolution ratio. However, the proposed method proved competitive, outperforming the uniform baseline. In retrievability, the proposed method achieved the highest score, surpassing both uniform and Top-K. This demonstrates that the adaptive selection strategy captures the most representative visual features of each identity, ensuring that the summary serves as a reliable query source for downstream re-identification. The lower score for Top-K suggests that selecting only the best frames can omit critical appearance variations needed for comprehensive matching.

The largest performance differences appear in the diversity metrics. Uniform sampling naturally yields high temporal coverage but exhibits redundancy, as indicated by a low visual diversity score. Top-K exhibits severe redundancy, producing the lowest visual diversity and a temporal coverage that tends to concentrate frames around a single high-quality segment. In contrast, our method achieves the best balance, with the highest visual diversity and temporal coverage, distributing keyframes across the entire trajectory and minimizing redundant frames.

Together, these results show that, while simplistic strategies such as Top-K or uniform sampling maximize specific individual metrics, they fail to provide a holistic summary. The proposed method balances biometric quality and narrative coverage, creating a concise and diverse summary optimized for both human interpretation and automated analysis.

**Table 7: Evaluation of VS on Custom-YT. We compare our proposed identity-centric approach against the Uniform and Top-K selection baselines across three evaluation categories.**

| Method | Optimization | | Information | Diversity | |
|---|---|---|---|---|---|
| | Biometric confidence | Resolution ratio | Retrievability | Visual | Temporal |
| Uniform | 0.7362 | 0.2381 | 0.9648 | 0.0832 | 0.8688 |
| Top-K | **0.8138** | **0.2524** | 0.9333 | 0.0370 | 0.4712 |
| Proposed | 0.7637 | 0.2438 | **0.9663** | **0.0909** | **0.8921** |

## 4.6. Qualitative analysis of summary selection

We performed a qualitative analysis using video sequences from the CHAD dataset (Danesh Pazho et al., 2023). Figure 3 presents the temporal distribution of keyframes for two distinct identities (ID 1 and ID 2).

The shaded area shows the continuous evolution of the importance score $S(P,t)$, while the markers denote the frames extracted by each selection strategy.

The Top-K strategy (green triangles) clusters its selections around the highest absolute peaks of the importance curve. While this guarantees high biometric quality, it produces severe temporal redundancy, capturing the subject within a very limited time span and completely missing the rest of their trajectory. In contrast, Uniform sampling (blue crosses) ensures temporal spread but is independent of the semantic content of the scene. As the plots show, uniform selections frequently fall into low-importance "valleys," extracting suboptimal frames where the subject may be occluded, distant, or blurred, simultaneously missing critical high-scoring events.

In contrast, the proposed method (red dots), driven by Adaptive Non-Maximum Suppression, achieves an optimal balance. It successfully anchors its selections to the semantic peaks of the sequence, capturing high-quality visual moments, while applying a temporal penalty to avoid redundancy. As the sequence illustrates, this ensures that the final summary captures a diverse and representative visual narrative of the subject's entire presence in the scene.

***Summarization guided by semantic queries:*** In addition, our framework is not limited to a single, rigid summary output. As illustrated by the radar chart in Figure 3 (right), the proposed scoring function enables dynamic, query-guided summarization through the use of Semantic Profiles. By adjusting the weight distribution ($w_1, w_2, w_3, w_4$), users can tailor the summary to specific observation intents. For example, an operator performing a "Forensic Search" can maximize the biometric and resolution weights to extract only the sharpest facial crops. Alternatively, selecting a "Social Dynamics" or "Action" profile shifts the algorithmic focus toward behavior, prioritizing frames where subjects interact with others or exhibit significant motion, even if their faces are partially obscured. This flexibility allows the system to adapt smoothly to diverse analytical requirements.

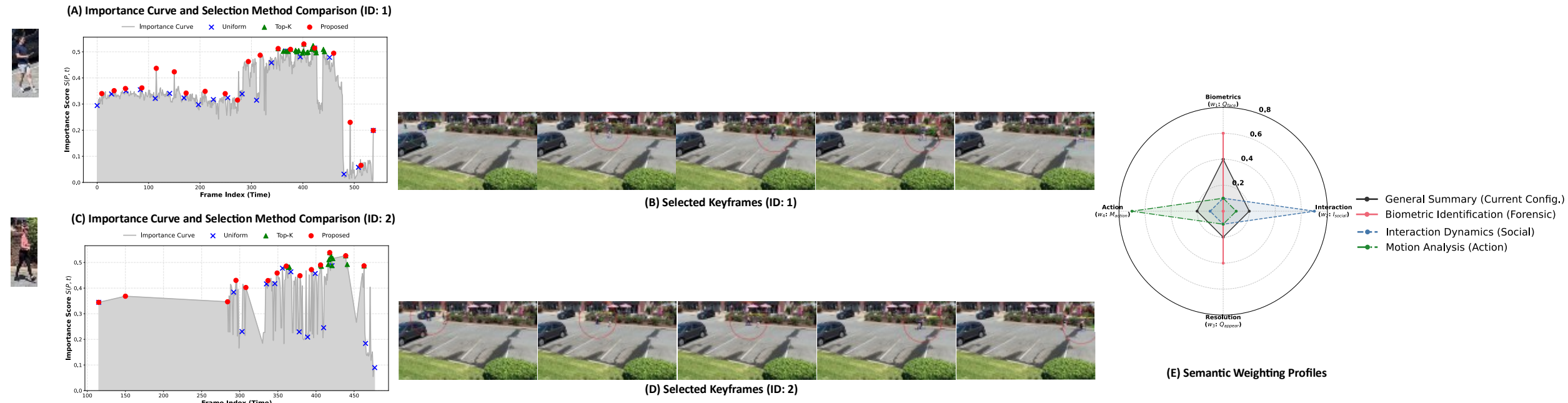

**Figure 3: Qualitative evaluation of summary selection strategies for two distinct identities. (A, C) Temporal evolution of the importance score S(P,t) for ID 1 and ID 2, respectively. Our proposed method (red circles) effectively captures semantically rich frames while maintaining temporal diversity, avoiding the severe redundancy of the Top-K baseline (green triangles) and the suboptimal semantic performance of Uniform sampling (blue crosses). (B, D) The resulting visual summaries generated by the proposed method, demonstrating a complete and diverse narrative sequence for each identity. (E) Semantic profiles illustrating the system's ability to generate query-guided summaries, enabling the dynamic creation of personalized summaries based on specific analytical intents.**

## 4.7. Failure analysis and limitations

Although the proposal demonstrates notable robustness to occlusion and appearance variation through 3D structural and fine-grained visual features, extreme stress scenarios reveal specific vulnerabilities. We detail below the most frequent failure modes, as well as the main limitation of the architecture. These factors are tied to the physical limits of the descriptors, the constraints of our temporal association model, and the processing load.

***Prolonged occlusion with non-linear motion:*** The system struggles with prolonged occlusions if the subject changes trajectory. Because spatial tracking relies on simple kinematics, drastic changes in motion behind an obstacle cause the location prediction to fail when the subject reappears, fragmenting the trajectory.

***Spatial overlap and feature entanglement:*** In dense crowds with overlapping bounding boxes, visual features become corrupted. This reduces the discriminative power of the network, introduces bias into the dynamic cost calculations, and can force a tracking failure.

***Image degradation:*** Extreme lighting degradation or severe motion blur directly impacts feature quality. In addition, 3D shape estimation is vulnerable to scale ambiguity. For very distant subjects, small bounding boxes lack detail, causing the 4D-Humans network to generate a noisy or generic mesh, temporarily neutralizing the advantage of our structural cues.

***Error propagation to the summary:*** If an identity switch is unrecoverable, the error contaminates the final summary, introducing several different people into the narrative of a single subject.

***Real-time processing:*** Despite the pre-filtering steps, the main limitation of the proposal is its computational cost, due to the expensive extraction of descriptors.

## 5. Conclusions

This work presents an extension of a previous identity-centric video summarization proposal. We incorporate a hierarchical association strategy based on biometric anchors, together with a feature set that integrates adaptive facial biometrics, Transformer-based appearance, and 3D body shape. This combination maintains identity consistency in scenarios with occlusion, pose changes, and visual ambiguity, producing summaries that are more coherent at the narrative level.

The results show that the hierarchical fusion of complementary signals reduces trajectory fragmentation and eliminates identity switches on the evaluated dataset, improving the overall stability of tracking. The proposed summarization module balances biometric quality, social context, and temporal diversity, generating compact representations that preserve the relevant information of each individual.

***Future work:*** The results and observations of this study open several priority lines of research:

- **Dataset expansion and large-scale validation:** While the presented evaluation acts as a rigorous stress test, it represents a bounded scenario. As a first line of work, we will expand the dataset to increase representativeness with respect to real-world diversity. We will also scale the evaluation to public, high-density benchmarks.
- **Semantic activity classification (HAR):** To enrich the summary scoring function, we propose integrating a human action recognition module. This will make it possible to distinguish between routine and anomalous events, an improvement that is especially relevant in video surveillance scenarios, where it will allow sequences of higher contextual value to be prioritized.
- **Optimization for real-time processing:** To enable online deployment without sacrificing accuracy, we will investigate architectural optimization and quantization techniques that accelerate inference.
- **Subjective perceptual evaluation:** Finally, since the narrative quality of a summary has an inherently subjective component that conventional objective metrics do not fully capture, we will design a perceptual evaluation framework with independent human users. This cross-study protocol will allow the summaries generated automatically by our architecture to be compared against manual selections created by participants, to formally evaluate the cognitive and semantic quality of the results using standardized qualitative assessment instruments and user perception scales.

## Acknowledgements

This work has been funded by the Recovery, Transformation, and Resilience Plan, financed by the European Union (Next Generation), through the LUCIA project (Fight against Cybercrime through the application of Artificial Intelligence), granted by INCIBE to the Universidad de León.